\documentclass[letterpaper, 10 pt, conference]{ieeeconf}  

\usepackage{amsmath} 
\usepackage{amssymb}  
\usepackage{glossaries}
\usepackage{cite}
\usepackage{orcidlink}
\let\labelindent\relax
\usepackage[inline]{enumitem}
\usepackage{multirow}
\usepackage{booktabs}
\usepackage[font=footnotesize]{caption}
\usepackage{subcaption}
\usepackage{svg}
\usepackage{algpseudocode}
\usepackage{algorithmicx}
\usepackage[linesnumbered, ruled, vlined]{algorithm2e}
\usepackage{graphicx}
\usepackage{tabularx}
\usepackage{balance}
\usepackage{wrapfig}

\SetKwInput{KwData}{Input}
\SetKwInput{KwResult}{Output}

\newcommand\gymrta[0]{\textit{GyMRTA}}

\title{\LARGE \bf
Learning Policies for Dynamic Coalition Formation \\
in Multi-Robot Task Allocation
}

\author{Lucas C. D. Bezerra~\orcidlink{0000-0002-3967-4374}, Ataíde M. G. dos Santos~\orcidlink{0000-0003-2725-1734}, and Shinkyu Park~\orcidlink{0000-0002-8643-404X}
\thanks{The work was supported by funding from King Abdullah University of Science and Technology (KAUST).}
\thanks{Bezerra and Park are with Electrical and Computer Engineering, King Abdullah University of Science and Technology (KAUST), Thuwal, 23955-6900, Kingdom of Saudi Arabia. {\tt \{lucas.camaradantasbezerra, shinkyu.park\}@kaust.edu.sa}}
\thanks{Santos is with the Department of Electrical Engineering, Federal University of Sergipe (UFS), São Cristóvão, Sergipe, 49107-230, Brazil. {\tt ataidegualberto.eng@gmail.com}}}

\newacronym{decpomdp}{Dec-POMDP}{Decentralized Partially-Observable Markov Decision Process}
\newacronym{rl}{RL}{Reinforcement Learning}
\newacronym{cnn}{CNN}{Convolutional Neural Network}
\newacronym{nn}{NN}{Neural Network}
\newacronym{marl}{MARL}{Multi-Agent Reinforcement Learning}
\newacronym{mrta}{MRTA}{Multi-Robot Task Allocation}
\newacronym{ctde}{CTDE}{Centralized Training and Decentralized Execution}
\newacronym{pg}{PG}{Policy Gradient}
\newacronym{vd}{VD}{Value Decomposition}
\newacronym{lbf}{LBF}{Level-Based Foraging}
\newacronym{stmrta}{ST-MR-TA}{Single-Task robots, Multi-Robot tasks, Time-extended Assignment}
\newacronym{cta}{CTA}{Cooperative Task Allocation}
\newacronym{mappo}{MAPPO}{Multi-Agent Proximal Policy Optimization}
\newacronym{iql}{IQL}{Independent Q-Learning}

\newacronym{kaust}{KAUST}{King Abdullah University of Science and Technology}
\newacronym{rv}{RV}{Random Variable}
\newacronym{iid}{i.i.d.}{independent and identically distributed}
\newacronym{mse}{MSE}{Mean Squared Error}
\newacronym{mpc}{MPC}{Model Predictive Controller}
\newacronym{pcfa}{PCFA}{Project manager-oriented Coalition Formation Algorithm}

\IEEEoverridecommandlockouts                              

\begin{document}

\maketitle

\begin{abstract}

We propose a decentralized, learning-based framework for dynamic coalition formation in \acrfull{mrta}. Our approach extends MAPPO by integrating spatial action maps, robot motion planning, intention sharing, and task allocation revision to enable effective and adaptive coalition formation. Extensive simulation studies confirm the effectiveness of our model, enabling each robot to rely solely on local information to learn timely revisions of task selections and form coalitions with other robots to complete collaborative tasks.  The results also highlight the proposed framework's ability to handle large robot populations and adapt to scenarios with diverse task sets.

\end{abstract}

\section{Introduction}

\begin{figure}[t]
    \centering
    \includegraphics[width=.8\linewidth]{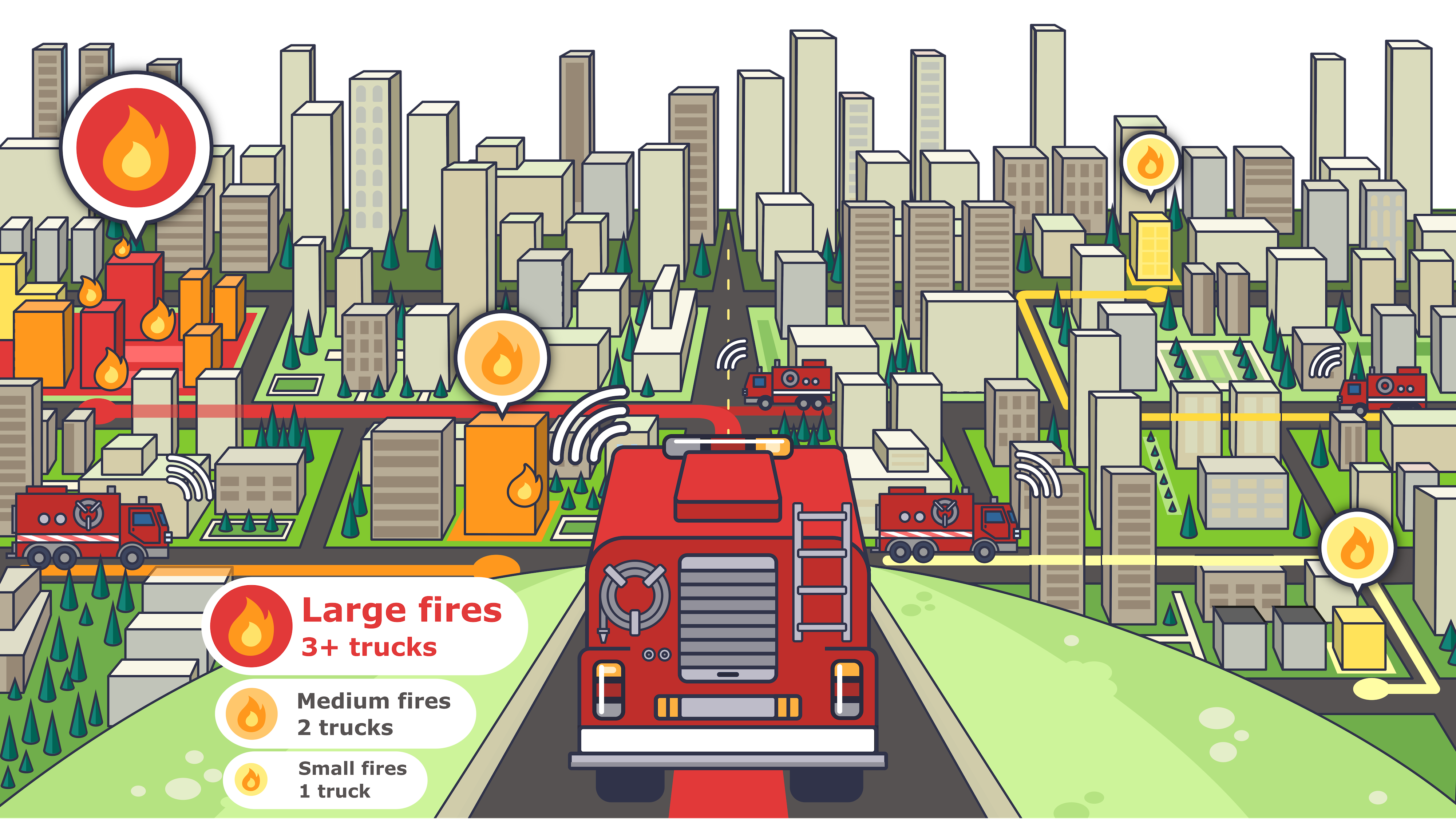}
    \caption{
    The firetruck dispatch as an illustrative example of a dynamic coalition formation problem: a team of firetrucks must form coalitions to extinguish fires that cannot be accomplished individually. In the event of a catastrophe that disrupts communication infrastructure, firetrucks may need to independently determine which fires to prioritize.}
    \label{fig:intro}
    \vspace{-1.0em}
\end{figure}

Coalition formation, widely observed in nature, such as in ant colonies and beehives, inspires researchers and engineers to design simple robots capable of strategic cooperation. In multi-robot systems, coalitions emerge when the collective actions of robots exceed what they could accomplish individually, as illustrated in Fig.~\ref{fig:intro}. In this work, we focus on developing policies for a team of robots capable of performing tasks that require coalition in dynamic environments. In such scenarios, robots must learn to coordinate by leveraging local information gathered from their own sensors and shared data from neighboring robots to effectively execute tasks. 

\gls{mrta} is a research theme directly relevant to our work, focusing on the challenge of assigning and executing multiple tasks across a team of robots to optimize a collective objective over an extended time horizon. We focus on MRTA problems where: 
\begin{enumerate*}[label=(\roman*)]
  \item each robot is capable of executing only one task at a time,
  \item tasks may require multiple robots to collaborate simultaneously for completion,
  \item robots must be adjacent to a task to execute it, necessitating movement to the task location beforehand, and
  \item completed tasks are removed, and new tasks continuously spawn in the environment.
\end{enumerate*}
This problem formulation falls under the \gls{stmrta} category of \gls{mrta}, as defined by the taxonomy in \cite{mrta_taxonomy}. Relevant applications include transportation \cite{Tuci2018} and 
service dispatch scenarios \cite{Mouradian2017, Couceiro2019} such as firefighting and large-scale disaster response.

We focus on developing a decentralized solution where each robot observes only its immediate surroundings. With such partial observability in \gls{mrta}, robots only identify tasks within their sensor range, causing the set of perceived tasks to change as they move. Additionally, we consider dynamic environments where new tasks continuously emerge over time. Well-established approaches to partially observable task allocation problems rely on inter-robot communication, where robots share their observations, internal state, and decision with their neighbors. Market-based approaches \cite{mrta2015, marketbased_survey} are among the most popular frameworks due to their robustness and scalability. However, these methods often rely on high-frequency communication between robots to reach agreement before task assignments are made \cite{marketbased_survey, Choi2009, pcfa}. Although this strategy is efficient in some restricted cases, it becomes impractical in dynamic environments.

We propose a learning-based framework for multi-robot dynamic coalition formation to address problems within the \gls{stmrta} \gls{mrta} category. In this framework, each robot utilizes a receding horizon-type planning strategy to generate a long-term plan for task execution, which is repeatedly revised as necessary in subsequent time steps. Each robot shares its plan with other robots that are within its communication range. This combination of task reassessment and sharing of the task execution plan enables dynamic coalition formation. Our contributions are summarized as follows:
\begin{enumerate}[label=(\roman*)]
    \item To the best of our knowledge and based on a recent survey article \cite{new_survey}, the problem of decentralized dynamic coalition formation under partial observability has not been previously addressed. To facilitate coalition, we extend a \gls{marl} algorithm by integrating 
    spatial action maps, robot motion planning, intention sharing, and task allocation revision. Compared to existing MARL approaches, such as \cite{formic, action_map, intention_map}, the integration of these key components fosters coalition formation through decentralized decision-making in dynamic environments (see Table~\ref{tab:model_comparison} for a comprehensive comparison).
    
    \item Each robot's policy is modeled as an end-to-end convolutional neural network based on the U-Net architecture \cite{unet}. The network processes information on nearby tasks and other robots within the robot's sensing range and integrates the intentions of neighboring robots. It is trained using the \gls{mappo} algorithm with the parameter-sharing technique.
        
    \item Through extensive simulations, we demonstrate that our approach outperforms existing methods including a market-based approach, \gls{pcfa}. We also demonstrate its scalability with up to 1000 robots and its generalizability across a wide range of task settings.
\end{enumerate}

This paper is organized as follows: \S\ref{sec:related_work} reviews related work in \gls{mrta} and \gls{marl}. In \S\ref{sec:problem_formulation}, we formally define the problem of dynamic coalition formation as a \acrshort{decpomdp}. \S\ref{sec:method} outlines our proposed method as a solution to the problem. In \S\ref{sec:exp}, we evaluate the effectiveness, scalability, and generalizability of the proposed method through extensive simulations, comparing it to the baseline. \S\ref{sec:conclusions} summarizes the paper and outlines future research directions.

\section{Related Work}
\label{sec:related_work}

\gls{mrta} is widely researched topic at the intersection of robotics and multi-agent systems \cite{rev1, rev2, Park2021, Park2023_1, Park2023_2, Notomista2019, rev3, trait, cocoa, marketbased_survey, pcfa, Xie2018, Choi2009}. Among the various problem settings, \gls{stmrta} \gls{mrta} formulations are most relevant to our work. 
Solutions to \gls{stmrta} \gls{mrta} include optimization-based \cite{Park2021, Park2023_1, Park2023_2, Notomista2019}, trait-based \cite{trait, cocoa}, and market-based \cite{marketbased_survey, pcfa, Xie2018, Choi2009} methods. Among decentralized solutions for \gls{mrta}, market-based approaches are particularly popular. Specifically, \acrshort{pcfa} \cite{pcfa} is a market-based solution designed for coalition formation in homogeneous teams with dynamic tasks. Due to the similarity in problem formulation, we selected \acrshort{pcfa} as the baseline for our simulations reported in \S\ref{sec:exp}.

\gls{rl} focuses on learning a policy (i.e., a mapping from observation to action) that maximizes a robot's total reward in a Markov decision process. It has been widely applied to various robotics problems, particularly in object manipulation \cite{action_map, relmogen, Sun2021, rtaw}. While many approaches aim to directly control robot actuators, studies such as \cite{action_map, relmogen, warehouse} propose the use of an abstract action map. In this framework, the policy selects a task at each time step, while a motion planner handles the low-level actuator control to navigate the robot to the task location. This abstraction alleviates the learned policy from handling low-level control, allowing it to concentrate on long-term planning. 

This approach is extended to the multi-robot domain in \cite{intention_map}, where robots not only select task locations but also communicate their chosen task locations to nearby robots. The authors show that this additional communication improves cooperation and enhances overall performance. However, their method does not address coalition formation. As demonstrated in \S\ref{sec:exp}, their approach, with appropriate adaptations, performs suboptimally compared to ours in scenarios that require effective coalition formation.
\gls{marl} naturally extends the conventional \gls{rl} framework to cooperative multi-robot scenarios, where instead of learning policies to maximize individual rewards, robots learn to maximize a global reward through coordination \cite{marl_book}. Several studies have adapted \gls{rl} algorithms for the multi-robot domain \cite{Foerster2017, mappo, lbforaging, qmix}. A widely adopted approach is the \gls{ctde} paradigm, where robots are trained using shared information \cite{mappo, lbforaging, qmix}. \gls{ctde} offers faster training and addresses non-stationarity.

Several studies have applied \gls{marl} to \gls{mrta} \cite{lbforaging, Shibata2022, formic, intention_map}. For example, \cite{lbforaging, Shibata2022} specifically address \gls{mrta} problems involving multi-robot tasks. The \gls{mrta} problem was introduced to the \gls{marl} community as a benchmark in \cite{lbforaging}, where the authors released a Python package called \gls{lbf}. In \gls{lbf}, each object is assigned a level, and it can only be picked up by a group of robots if the sum of the robots' levels meets or exceeds the object's level. Although \gls{lbf} has become a popular \gls{marl} benchmark \cite{epymarl, Cao2023, Boggess2022}, it lacks support for dynamic object spawning and abstract action spaces (e.g., selecting task positions on the grid). In \cite{epymarl}, several \gls{marl} algorithms were evaluated across popular benchmarks, including \gls{lbf} (in fully observable scenarios only), and the EPyMARL package was released for community use. We expand their implementation by incorporating robot communication and additional features, which are detailed further in \S\ref{sec:method}.

In \cite{Shibata2022, formic}, \gls{marl} is applied to an extended version of \gls{mrta}, where robots must not only collect objects but also transport them. In \cite{Shibata2022}, which involves multi-robot tasks, the authors propose a novel inter-robot communication architecture. However, their approach assumes a fully observable environment. In \cite{formic}, the environment is defined using a multi-channel, grid-like observation space under partial observability, with the robots' policy modeled using a convolutional neural network, similar to our approach. However, unlike our work, they do not employ an abstract action space and use a shallower policy architecture. 

\begin{figure*}[t]
\centering
    \centering
    \includegraphics[width=.9\linewidth]{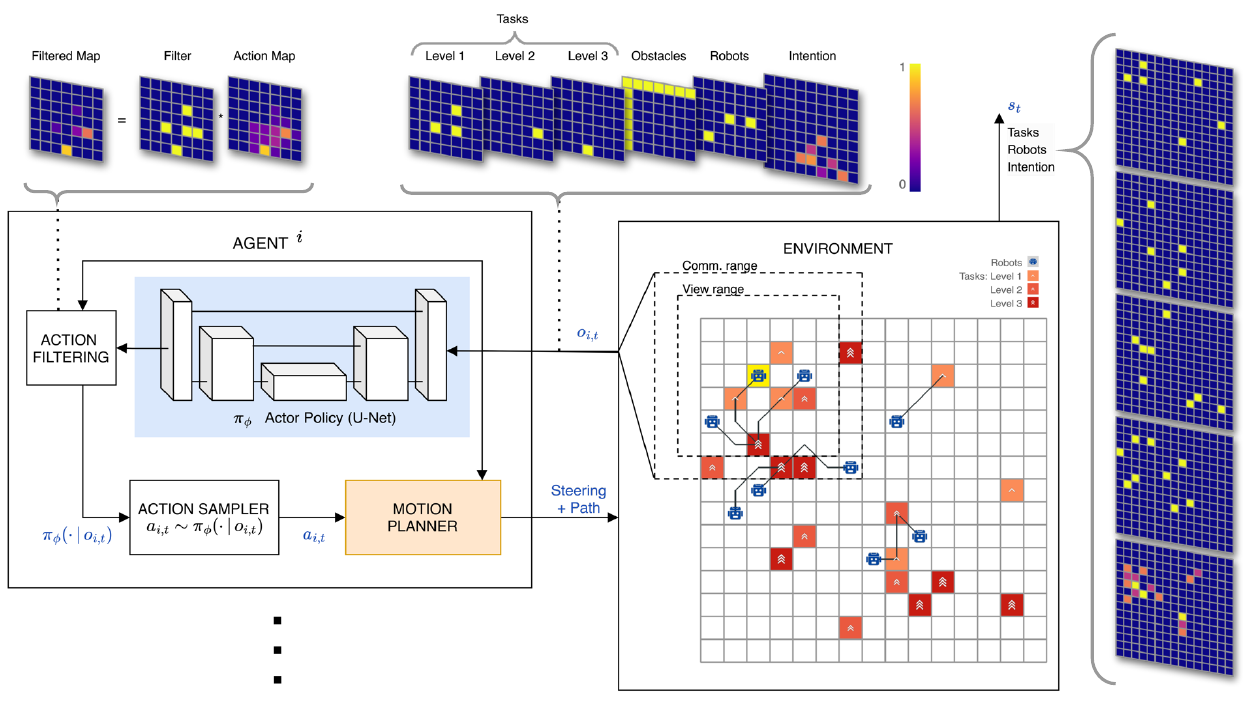} 
    
\caption{MRTA Environment-Robot Interaction: On the right, a depiction of the environment is shown along with the state representation $s_t$ used by the critic during training. Each robot receives an observation $o_{i,t}$ consisting of multiple channels. The observation is then fed into the policy $\pi_\phi$ and filtered to consider only visible tasks. A task is selected from the policy output, and the motion planner finds a feasible path to it.}
\label{fig:mappoDiagram}
\vspace{-1em}
\end{figure*}

\section{Problem Description}
\label{sec:problem_formulation}

Consider a rectangular 2D grid environment of size $W \times W$, bounded by walls on all sides, where robots and tasks are distributed, as illustrated in Fig.~\ref{fig:mappoDiagram}. Each cell $p$ in the grid belongs to the cell space $\mathbb{P}=\{1,\dots,W\} \times \{1,\dots,W\}$ and can be either empty or occupied by a robot, a task, or a wall. The grid contains a homogeneous team of $N$ robots, where the position of the $i$-th robot at time step $t$ is denoted as $p_{i,t} \in \mathbb{P}$. At each $t$, a robot can either remain in its current position or move to an adjacent cell, including diagonal moves. Robots observe the cells within their view range, defined by $R_{\text{view}}$ cells, and communicate with others within a communication range of $R_{\text{comm}}$ cells. 

Each task is associated with a positive integer $l_i \in \{1, \dots, L_{max}\}$, referred to as the task's \textit{level}, which represents the number of robots required to execute it. Here, $L_{max}$ denotes the maximum task level. We consider a dynamic task set: tasks are removed from the environment once completed by robots, and new tasks can appear at any time step in unoccupied cells. At time step $t$, the set of available tasks is represented as $\mathbb T_t = \{ ( q_{i},  l_i ) \}_{i=1}^{T_t}$, where $q_{i}$ denotes the location of the $i$-th task, $l_i$ indicates its level, and $T_t$ represents the total number of tasks available.

We consider two task spawn settings:
\begin{enumerate*}[label=(\roman*)]
    
    \item \textit{Bernoulli ($p$) spawn}: At each time step $t$, a new task can appear in any unoccupied cell with probability $p$. The level of each newly spawned task is sampled from a discrete uniform distribution $\mathcal{U}(\{1,\dots, L_{max}\})$.

    \item \textit{Instant respawn}: When a task is completed, a new task of the same level immediately respawns in one of the unoccupied cells. This setting maintains a constant task density throughout the episode.

\end{enumerate*}

To design a decentralized policy under partial observability, we model the problem as a \gls{decpomdp} \cite{introduction-decpomdps,decpomdp}. A \gls{decpomdp} is defined by the tuple $( N, \mathcal S, \mathcal A, \mathcal T, \mathcal O, O, R )$, where $N$ is the number of robots, $\mathcal S$ represents the state space of the environment, $\mathcal A$ and $\mathcal O$ denote the action and observation spaces of individual robots, respectively. The state transition function is defined as $\mathcal T: \mathcal{S}\times\mathcal{S}\times\mathcal{A}^N \to \mathbb{R}_+ $, and the reward function as $R: \mathcal{S} \times \mathcal{A}^N \to \mathbb{R}$.\footnote{Here, we provide a general description of a \gls{decpomdp}; implementation details specific to our framework are presented in \S\ref{sec:proposed_model}.}

Each robot $i$ receives an observation $o_{i,t} \in \mathcal{O}$ at time $t$ and independently selects an action $a_{i,t} \in \mathcal A$ based on its (stochastic) policy $\pi: \mathcal A \times \mathcal O \to \mathbb R_+$, i.e., $a_{i,t} \sim \pi(\cdot | o_{i,t})$. 
An observation $o_{i,t}$ is a function of the state $s_t \in \mathcal S$ and is defined as $o_{i,t} = O(s_t, i)$, where $O: \mathcal{S}\times\{1,\dots,N\} \to \mathcal{O}$ is the observation function. 
Given the state $s_t$ and the joint action $\mathbf{a}_t = (a_{1,t}, \dots, a_{N,t}) \in \mathcal{A}^N$, the next state $s_{t+1} \in \mathcal S$ is sampled according to the state transition function $\mathcal T$, i.e., $s_{t+1} \sim \mathcal T(\cdot | s_t, \mathbf a_t)$. The transition function $\mathcal T$ captures both deterministic dynamics (e.g., robot movement, task execution) and stochastic dynamics (e.g., task spawning) of the environment. For each state-action pair, a shared reward $r_t = R(s_t, \mathbf a_t)$ is computed for all robots.

In a \gls{decpomdp}, state-action trajectories are represented as an episode $\xi := ( s_0, \mathbf{a}_0, s_1, \mathbf{a}_1, \dots, s_T )$. Given an initial state distribution $P_0: \mathcal S \to \mathbb{R}_+$ and the episode likelihood $P_\pi(\xi) = P_0(s_0) \prod_{t=0}^{T-1} \prod_{i=0}^{N-1} \pi(a_{i,t}|o_{i,t}) \, \mathcal T(s_{t+1}|s_t,\mathbf{a}_t)$, the performance of a policy $\pi$ can be evaluated using the expected episodic reward $\mathbb E_{\xi \sim P_\pi(\cdot)} \big[ {\sum_{t=0}^T R(s_t, \mathbf{a}_t)} \big]$, where $T$ is the terminal time. When a sufficiently large number of episodes is available in a dataset $\mathcal D$, this expectation can be approximated as $\mathbb E_{\xi \sim \mathcal D} [ {\sum_{t=0}^T R(s_t, \mathbf{a}_t)} ]$, thereby removing the need for explicit knowledge of $P_0$ and $\mathcal T$.


\section{Task Allocation Policy for Dynamic Coalition Formation}
\label{sec:method}

We elaborate on the two main contributions of this work: \begin{enumerate*}[label=(\roman*)]
\item the design of decentralized task allocation policy structure, and 
\item the development of a learning-based algorithm within the Dec-POMDP framework to identify the optimal policy. 
\end{enumerate*}
Our proposed policies are designed to select tasks for execution, assess whether a robot’s current task allocation requires revision, and determine new task assignments based on the robot’s evolving local information. Once a task is assigned, each robot performs low-level control to move to a neighboring cell along a collision-free path within a single time step. In the subsequent time step, the policy reassesses the task allocation based on updated local information, redirecting the robot to a new task if necessary. This iterative process ensures that the overall collective objective is optimized.

\subsection{Dec-POMDP and Task Allocation Policy} \label{sec:proposed_model}
\subsubsection{Reward Function}
Higher-level tasks, which require coalitions of multiple robots, are more challenging to complete. To prevent robots from favoring lower-level tasks, we prioritize higher-level tasks by rewarding robots with a value equal to the square of the task level. Specifically, the reward at each time step $t$ is defined as $r_t = \sum_{(q_i, l_i) \in \mathbb T_t \backslash \mathbb T_{t+1}} l_i^2$, where $\mathbb T_t \backslash \mathbb T_{t+1}$ represents the set of tasks completed at time step $t$, and $l_i$ denotes the level of task~$i$.\footnote{Once the coalition requirement is met, we assume that the robots can complete each task within a single time step.}

\subsubsection{State and Observation Spaces}
We adopt a 2D occupancy grid representation segmented into multiple channels (see Table \ref{tab:channels}), enabling the use of \gls{cnn} architectures for policy design, similar to approaches in \cite{formic, action_map, intention_map}. This differs from the representation used in \gls{lbf} \cite{lbforaging}, which encodes the state as a vector of robots and their positions. In our framework, the state is defined as multiple separate channels as $s_t \triangleq (\mathcal{C}^{\text{robot}}_{t}, \mathcal{C}^{\text{task}}_t, \mathcal{C}^{\text{obs}}_t,  \mathcal{C}^{\text{intent}}_t)$, where $\mathcal{C}^{\text{intent}}_t = \mathcal{C}^{\text{intent}}_{1,t} \times \cdots \times \mathcal{C}^{\text{intent}}_{N,t}$.\footnote{To simplify the design and improve the efficiency of critic model training, our implementation consolidates $\mathcal{C}^{\text{intent}}_t$ into a single aggregated channel that combines the intention channels of all robots. Empirical studies indicate that this aggregation does not impact the training outcomes.}

The observation $O(s_t, i)$ for robot~$i$ is derived by projecting and cropping $s_t$ onto areas centered around the robot's position, determined by its view range $R_{\text{view}}$ and communication range $R_{\text{comm}}$. Specifically, the robot intention channel $\mathcal{C}^{\text{intent}}_{t}$ includes $\mathcal C^{\text{intent}}_{j,t}$ from robot~$j$ if $j$ is within $R_{\text{comm}}$ cells of robot~$i$. The other channels, $\mathcal{C}^{\text{robot}}_t, \mathcal{C}^{\text{task}}_t, \mathcal{C}^{\text{obs}}_t$, are obtained by cropping $s_t$ according to $R_{\text{view}}$. 

\begin{table}[t]
\scriptsize
\centering
\caption{Description of CNN input channels.}
\label{tab:channels}
\begin{tabular}{@{}lll@{}}
\toprule
\textbf{Notation} & \textbf{Space} & \textbf{Description} \\ \midrule
$\mathcal{C}^{\text{robot}}_t$ & $\{0, 1\}^{W \times W}$ & Robots \\
$\mathcal{C}^{\text{task}}_t$ & $\{0, 1\}^{W \times W \times L_{max}}$ & Tasks (separated by level) \\
$\mathcal{C}^{\text{obs}}_t$ & $\{0, 1\}^{W \times W}$ & Obstacles \\
$\mathcal{C}^{\text{intent}}_{i,t}$ & $\mathbb{R}_+^{W \times W}$ & Robot~$i$'s intention map \\
\bottomrule
\end{tabular}%
\vspace{-1.7em}
\end{table}

\subsubsection{Action Space, Motion Planner, and Intention Map}
\label{sec:motion_planner}

Following a similar representation as in \cite{action_map, relmogen}, a robot's action -- determined by the task allocation policy $\pi_\phi$ -- is defined as the location of its next task. Robots can select task locations only if they are within its view range; if no tasks are within view range, all grid locations within communication range are available for action selection. This constraint provides prior information, reducing training time while encouraging robots to commit to specific tasks. 

After selecting a task location, a motion planner computes the shortest collision-free path to that location while avoiding other robots, obstacles, and other tasks. We adopt the A* search algorithm for simplicity. The path is dynamically recomputed whenever a potential collision is detected. Robots follow these updated paths to their assigned task locations. To improve coalition formation for higher-level tasks, the policy $\pi_\phi$ assigns actions by considering the long-term task execution plans of neighboring robots. 

\begin{algorithm}[t]
\small
\caption{\small Robot Motion Control} \label{alg:robot_motion_control}
    \DontPrintSemicolon
    \SetKwFunction{FMotionPlanning}{MotionPlanning}
    \SetKwProg{Fn}{Function}{:}{}
    \SetKwFunction{proc}{NextRobotPosition}
    \SetKwProg{myproc}{Procedure}{:}{}

    \Fn{\FMotionPlanning{$p_{i,t}, a_{i,t}$}} {
        Using the A* algorithm, find a collision-free path $(\bar p_1, \cdots, \bar p_n)$ where $\bar p_1 = p_{i,t}$ and $\bar p_n = a_{i,t}$.
        
        \Return $(\bar p_1, \cdots, \bar p_n)$
        
    }
    \myproc{\proc{$p_{i,t}, o_{i,t}$}} {
    $a_{i,t} \sim \pi_\phi (\cdot | o_{i,t})$
    \tcp*[l]{\footnotesize{draw a task location using the learned policy $\pi_\phi$ with local observation $o_{i,t}$.}}

    $(\bar p_1, \cdots, \bar p_n)$ = \FMotionPlanning($p_{i,t}, a_{i,t}$)   

    Generate an intention map with $(\bar p_1, \cdots, \bar p_n)$ and share it with neighboring robots.

    $p_{i,t+1} \gets \bar p_2$ 
    \tcp*[l]{\footnotesize{assign $\bar p_2$ as the robot's next position to move to.}}
    
    \Return $p_{i,t+1}$
    }
\end{algorithm}

We also employ an \textit{intention map} \cite{intention_map} that encodes each robot’s task location and collision-free path. Inspired by the encoding scheme in \cite{intention_map}, we use an exponentially decaying path representation, where the $i$-th position on the path is weighted by $\alpha^{-i}$, with $\alpha = 2/3$ in our simulation, and embedded in the intention map. This intention map is shared among neighboring robots to facilitate coalition formation. 
Algorithm~\ref{alg:robot_motion_control} outlines the motion control process for each robot guided by the learned policy $\pi_\phi$.

\subsection{Optimal Task Allocation Policy}
\label{sec:target_revision}

\subsubsection{\gls{mappo}}

To compute the optimal policy, we adopt \gls{mappo} \cite{mappo}, a \gls{ctde} algorithm designed for MARL. \gls{mappo} enables robots to share experiences during the learning phase, which is particularly beneficial for training policies in teams of homogeneous robots. MAPPO is an Actor-Critic algorithm comprising two distinct networks: the actor $\pi_\phi$ and the critic $V_\theta$. The actor outputs a distribution over actions based on each robot's observations, formally defined by the policy $\pi_\phi: \mathcal{A} \times \mathcal{O} \to \mathbb{R}_+$, where $\phi$ represents the parameters of the policy network. These parameters $\phi$ are shared across all robots, a technique known as \textit{parameter sharing} \cite{parameter_sharing}, which improves training efficiency, especially in cooperative scenarios involving homogeneous robots. The critic computes an approximate value function $V_\theta: \mathcal{S} \to \mathbb{R}$, parameterized by $\theta$, to estimate the expected discounted sum of future rewards $\mathbb E [ {\sum_{\tau=t}^T \gamma^{\tau-t} R(s_\tau, \mathbf{a}_\tau)} ]$.

The critic $V_\theta$ is trained to accurately estimate future rewards, while the actor $\pi_\phi$ is trained to select actions that maximize the critic's estimate. In our framework, both the actor $\pi_\phi$ and the critic $V_\theta$ are implemented using U-Nets \cite{unet}, a \gls{cnn}-based architecture. For this implementation, the actor's input (observations) and output (action distribution) as well as the critic's input (state), are represented as images (3D arrays). The U-Net architecture is particularly advantageous due to its ability to capture both local spatial relationships and global context within the input image, making it well-suited for policy learning.

\begin{table*}[t]
\centering
\caption{}
\begin{subtable}[t]{0.28\linewidth}
\centering
\caption{\gls{mappo} Parameters.}
\resizebox{\linewidth}{!}{
\begin{tabular}{@{}lll@{}}
\toprule
\textbf{Parameter} & \textbf{Value} & \textbf{Description} \\ \midrule
$\eta$ & $5 \cdot 10^{-4}$ & Learning rate \\
$\gamma$ & 0.95 & Discount factor \\
$T_{ret}$ & 5 & Adv. estimation steps \\
$\varepsilon$-clip & 0.2 & Clipping in PPO loss \\
$c_s$ & 0.01 & Entropy coefficient \\
Batch size & 10 & \\
Buffer size & 10 & \\
Reward norm. & Yes & \\
Return norm. & No & \\ 
\end{tabular}
}
\label{tab:mappo_settings}
\end{subtable}\hfill
\begin{subtable}[t]{0.33\linewidth}
\centering
\caption{Default environment settings.}
\resizebox{\linewidth}{!}{
\begin{tabular}{@{}lcc@{}}
\toprule
\textbf{Parameter}                                                   & \multicolumn{2}{c}{\textbf{Value(s)}}                                                                                 \\ \midrule
Grid size                                                   & \multicolumn{2}{c}{20 x 20}                                                                                           \\
View range                                                  & \multicolumn{2}{c}{5}                                                                                                 \\
Comm. range                                                 & \multicolumn{2}{c}{8}                                                                                                 \\
$L_{max}$                                                 & \multicolumn{2}{c}{3}                                                                                                 \\
Task spawn                                                  & \multicolumn{2}{c}{\begin{tabular}[c]{@{}c@{}}Instant respawn / \\ Bernoulli ($p=\{0.01,\dots,0.04\}$)\end{tabular}} \\
Task distribution                                           & Random corner patch                                             & Homogeneous                                             \\ \cmidrule(l){2-3} 
Number of agents                                            & 10                                                          & 40                                                      \\
\begin{tabular}[c]{@{}l@{}}Initial task setting \\ (see Table \ref{tab:gen_levels}) \end{tabular} & M2                                                          & 4xM2                                                    \\ 
\end{tabular}
}
\label{tab:main_exp_settings}
\end{subtable}\hfill
\begin{subtable}[t]{0.38\linewidth}
\centering
\caption{Model definitions and corresponding work.}
\resizebox{\linewidth}{!}{
\begin{tabular}{@{}cccccc@{}}
\toprule
\multirow{2}{*}{\textbf{Name}} & \multicolumn{4}{c}{\textbf{Components}} & \multirow{2}{*}{\textbf{\begin{tabular}[c]{@{}c@{}}Related \\ Work\end{tabular}}} \\ \cmidrule(lr){2-5}
 & \textbf{\begin{tabular}[c]{@{}c@{}}Action\\ Filtering\end{tabular}} & \textbf{\begin{tabular}[c]{@{}c@{}}Spatial\\ Actions\end{tabular}} & \textbf{\begin{tabular}[c]{@{}c@{}}Intention\\ Sharing\end{tabular}} & \textbf{\begin{tabular}[c]{@{}c@{}}Task\\ Revision\end{tabular}} &  \\ \midrule
PCFA & \checkmark &  & \checkmark & \checkmark & \cite{pcfa} \\
I & \checkmark &  &  &  & \cite{formic} \\
II & \checkmark & \checkmark &  &  & \cite{action_map} \\
III & \checkmark & \checkmark & \checkmark &  & \cite{intention_map} \\
\textbf{Ours} & \checkmark & \checkmark & \checkmark & \checkmark & -- \\ 
\end{tabular}
}
\label{tab:model_comparison}
\end{subtable}
\noindent\rule{\linewidth}{1pt}
\vspace{-1.0em}
\end{table*}

\subsubsection{Learning Task Allocation and Revision}
As shown in Fig.~\ref{fig:mappoDiagram}, we design a task allocation policy that assigns a task location $a_{i,t}$ to each robot based on its local information: $a_{i,t} \sim \pi_\phi (\cdot | o_{i,t})$, where $o_{i,t} = O(s_t, i)$. In a partially observable environment with dynamically spawning tasks, a robot may encounter new tasks and other robots as it moves, which were previously unaccounted for. This dynamic nature makes it challenging for multiple robots to form coalitions for higher-level tasks. Consequently, the policy must allow robots to revise their task execution plans when necessary. However, frequent task revisions can disrupt coalition formation and hinder the completion of higher-level tasks.

To address this challenge, we extend the \gls{mappo} framework to train a policy capable of both task allocation and revision. Unlike existing approaches \cite{intention_map, warehouse}, where task revisions occur only after a robot reaches its assigned task location and either completes it or fails due to the absence of a coalition, our trained policy proactively reallocates tasks as needed, based on evolving local observations and information shared by neighboring robots. As demonstrated in \S\ref{sec:exp}, this integrated approach enables our framework to concurrently learn task allocation and revision, ensuring optimization of the overall objective in the \gls{decpomdp} environment.

\section{Evaluation}
\label{sec:exp}

We conducted a series of simulations to evaluate:
\begin{enumerate*}[label=(\roman*)]
\item the performance of our proposed framework compared to existing methods, including the state-of-the-art market-based approach \gls{pcfa},\footnote{The original \gls{pcfa} \cite{pcfa} was modified to accommodate robots with limited view and communication ranges and to enable task reassignment. Details of these modifications are provided in the appendix.} 
\item its scalability with respect to the number of robots, and 
\item its generalizability across a range of task allocation scenarios with varying task difficulties.
\end{enumerate*}
Furthermore, we present an ablation study to underscore the critical role of task revision in dynamic coalition formation.

\subsection{Simulation Setup and Performance Metric}

To evaluate our proposed method, we developed \gymrta, a Gym environment for discrete-space, discrete-time \gls{mrta}, and extended the existing \gls{mappo} implementation from the EPyMARL \cite{epymarl} package. Our extensions include features such as task revision, action filtering, multi-dimensional observation and action spaces, support for custom state space, and tailored actor and critic \gls{nn} architectures. 
The hyperparameters of \gls{mappo}, presented in Table~\ref{tab:mappo_settings}, remain constant across all simulations. Each episode is limited to 100 time steps ($T=99$) to ensure a finite episodic reward.

At the start of each episode, the initial positions of robots and tasks are randomly sampled. Table \ref{tab:gen_levels} details on the initial number of tasks per level across various task settings used in the simulations. The environment is categorized based on task distribution as either homogeneous or non-homogeneous.

\begin{table}
\centering
\caption{Task settings.}
\label{tab:gen_levels}
\scriptsize
\begin{tabular}{@{}llllllll@{}}
\toprule
\multirow{2}{*}{
\begin{tabular}[c]{@{}l@{}}
\textbf{Setting} \\ 
\textbf{Name}
\end{tabular}
} 
& \multicolumn{3}{c}{\textbf{Tasks}} 
& \multirow{2}{*}{
\begin{tabular}[c]{@{}l@{}}
\textbf{Setting} \\ 
\textbf{Name}
\end{tabular}
} 
& \multicolumn{3}{c}{\textbf{Tasks}} \\ 
\cmidrule(l){2-4} \cmidrule(l){6-8} 
& \textbf{1} & \textbf{2} & \textbf{3} 
& & \textbf{1} & \textbf{2} & \textbf{3} \\ \midrule
\textbf{Easy} &    &    &    & \textbf{Hard}  &    &    &    \\
E1            & 15 & -  & -  & H1             & -  & 5  & 5  \\
E2            & 10 & -  & -  & H2             & -  & -  & 10 \\
E3            & 5  & 5  & -  & H3             & -  & -  & 5  \\ \midrule
\textbf{Medium} &   &    &   & \textbf{Non-Homogeneous}   &    &    &    \\
M1             & 6  & 4  & 4  & M2           & 4  & 3  & 3  \\ 
M2             & 4  & 3  & 3  & \textbf{Homogeneous}   &    &    &    \\
M3             & 2  & 2  & 1  & 4$\times$M2           & 16  & 12  & 12  \\ \midrule
\end{tabular}
\vspace{-2em}
\end{table}


In the homogeneous scenario, tasks are evenly distributed across the grid, maintaining a uniform density. Conversely, the non-homogeneous scenario features an uneven distribution of tasks, leading to variations in task density. To define non-homogeneous configurations, we adopt the \textit{random corner patch distribution}: the grid is divided into four equal-sized corner patches, and at the start of each episode, one patch is randomly chosen to contain all tasks. Within the chosen patch, the task density is uniform, while the rest of the grid remains empty. These scenarios are used to evaluate whether the learned policy enables robots to form effective coalitions for task execution while efficiently locating tasks.  

The primary performance metric is the average episodic reward. We calculate the mean and standard deviation of the episodic reward across 480 episodes: 5 different seeds with 96 evaluation episodes per seed. Each seed initializes the policy and environment with different parameters ($\theta, \phi$, and initial locations of robots and tasks) to ensure results are not biased by any specific initialization.

\textit{Computational Resources:} The policies were trained for 4M steps over a period of up to 13 hours and up to 68 hours in the non-homogeneous and homogeneous settings, respectively. Each policy is trained using 16 CPU cores (Intel Xeon Platinum 8260) and one GPU (Nvidia V100). After training, however, each trained policy executes in just $2.1 \pm 1.5$ ms on a single CPU core of the same model.

\subsection{Simulation Results}

\begin{figure*}[t!]
    \centering
    \begin{subfigure}[]{.5\linewidth}
        \centering
        \includegraphics[width=\linewidth]{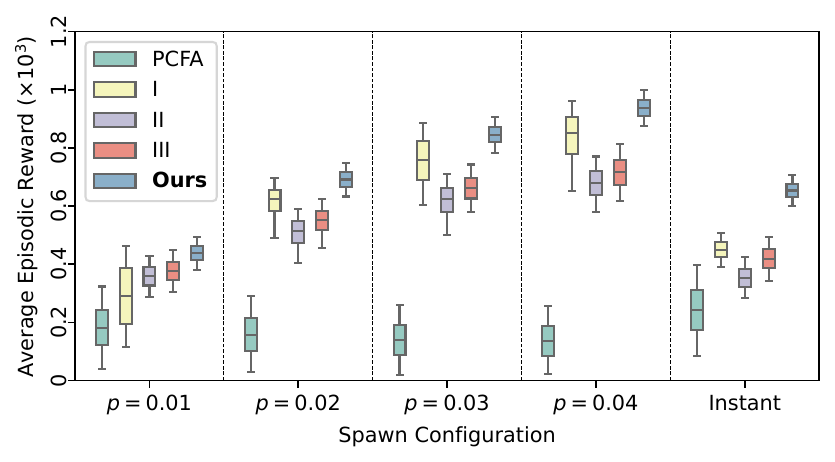}
        \caption{Non-homogeneous tasks.}
        \label{fig:boxplot3a}
    \end{subfigure}
    \begin{subfigure}[]{.49\linewidth}
        \centering
        \includegraphics[width=\linewidth]{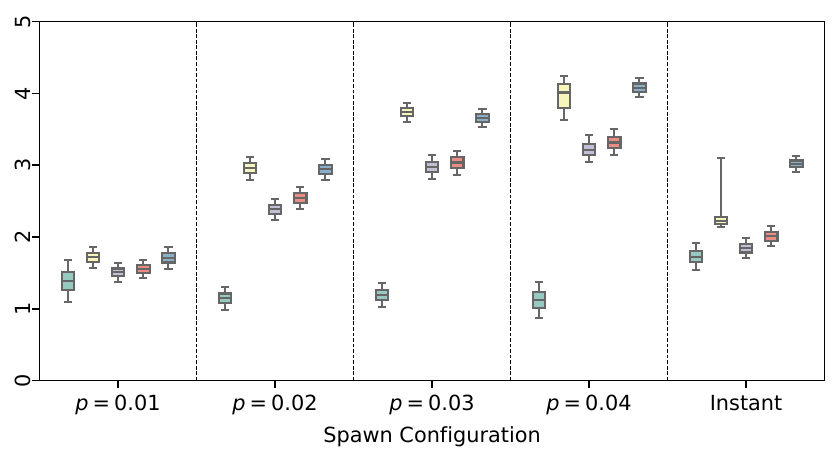}
        \caption{Homogeneous tasks.}
        \label{fig:boxplot3b}
    \end{subfigure}
    \caption{Setting-specific performance across various spawn configurations.}
    \label{fig:boxplotMain}
    \vspace{-1em}
\end{figure*}

\begin{figure}
    \centering
    \includegraphics[width=1.\linewidth]{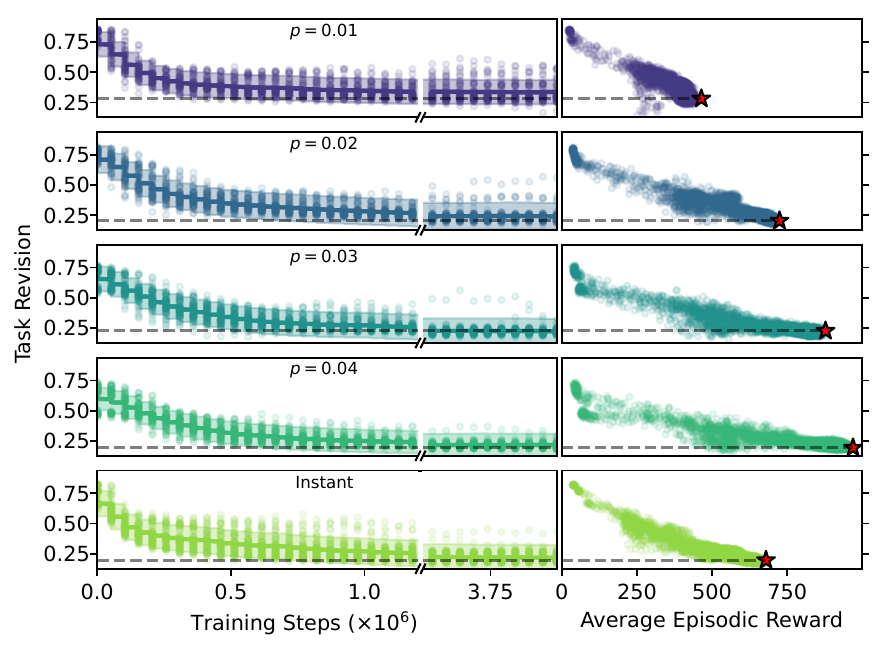}
    \caption{The graphs illustrate how task revision rates -- defined as the average number of time steps between each robot’s task selection changes -- evolve over the course of training, and how the average episodic reward improves as the robots learn optimal revision strategies. The dotted horizontal lines indicate the optimal revision rates.
    }
    \label{fig:task_rev}
    \vspace{-1.5em}
\end{figure}

We conducted a series of simulations to evaluate the proposed framework and the baseline across various environment settings, characterized by different task spawn configurations and distributions, while keeping other parameters constant. A summary of all simulation parameters is provided in Table \ref{tab:main_exp_settings}. In the homogeneous task distribution scenario, the task spawning area is four times larger than in the non-homogeneous case. To ensure consistent robot and task densities across both distributions, the number of robots and tasks in the homogeneous scenario is proportionally increased by a factor of four. For this purpose, we use the M2 and 4$\times$M2 settings for the non-homogeneous and homogeneous scenarios, respectively, as described in Table~\ref{tab:gen_levels}.

Fig.~\ref{fig:boxplotMain} compares the performance of our model, the baseline (PCFA), and various ablation studies on the components of our model across non-homogeneous and homogeneous task distributions and five task spawn configurations (Bernoulli spawn with probabilities $p =0.01, 0.02, 0.03, 0.04$ and instant respawn), spanning 10 different settings. Table \ref{tab:model_comparison} provides details of the models used in the ablation studies, highlighting their components and referencing relevant existing works associated with each model.\footnote{\textit{Spatial action} refers to defining the action in Dec-POMDP as the location of a task. Consequently, in model~I, which does not incorporate spatial action, an action is defined as a neighboring cell on the grid that a robot can move to within a single time step.} Each model was trained and evaluated in every environment setting.

Models I, II, and III in this study are closely related to existing works \cite{formic}, \cite{action_map}, and \cite{intention_map}, respectively, but include the following necessary modifications:
\begin{enumerate*}[label=(\roman*)]

    \item All three models are trained using \gls{mappo} within a \gls{decpomdp} framework, whereas the original references rely on the MDP assumption with individual rewards.
    
    \item While \cite{action_map} focuses on individual-robot training scenarios, model~II is trained for \gls{mrta} using our framework.
    
    \item In \cite{intention_map}, only level-1 tasks are considered, and a method similar to \gls{iql} \cite{iql}, which encourages competition rather than cooperation among multiple robots, is used, while model~III uses MAPPO.
\end{enumerate*}

As shown in Fig.~\ref{fig:boxplotMain}, our method consistently outperforms PCFA in all configurations. Notably, as the spawn rate $p$ increases, PCFA's performance declines, indicating its difficulty in coordinating robots to effectively prioritize and allocate tasks of varying difficulty levels. Moreover, we observe that task revision significantly improves performance, particularly in highly dynamic scenarios such as the instant respawn and Bernoulli spawn with $p=0.04$. These results align with our expectation that in environments where tasks spawn frequently and dynamically, robots must continuously reassess and update their long-term task execution plans. 

While frequent adjustments to task execution plans might seem disruptive to coalition formation among multiple robots working on high-level tasks, our proposed method enables robots to learn an optimal task revision strategy, resulting in superior performance. Fig.~\ref{fig:task_rev} depicts how the robots improve their average episodic reward by learning optimal task revision strategies. Importantly, this behavior is not manually encoded -- specifically, the task revision rate is not a learnable parameter -- but rather emerges from the learning process within our framework. Through repeated training, the robots progressively refine their optimal task revision rates, leading to more effective coalition formation and consistent improvements in average reward.

The ablation studies evaluate the impact of intention sharing and task revision on our model’s performance. Models without task revision are limited to selecting a new task only upon reaching assigned tasks, regardless of whether it executes the tasks or not. As shown in Fig.~\ref{fig:boxplotMain}, model~II (without intention sharing or task revision) underperforms both model~III (with intention sharing only) and our model (with both intention sharing and task revision), highlighting the importance of learning the optimal timing for task revision and its communication with neighboring robots.

In the homogeneous task distribution case, model~I performs comparably to our model across all Bernoulli spawn settings but underperforms in the instant respawn scenario. This outcome indicates that homogeneous task scenarios with Bernoulli spawn do not necessitate long-term planning. However, in the instant respawn scenario, the constant task density discourages simplistic approaches like those employed by model~I, resulting in its reduced performance.

\subsection{Scalability and Generalizability Evaluation}

To evaluate scalability, we first trained policies in a homogeneous task distribution environment ($W = 10$) with a fixed number of robots. The trained policy was then tested while progressively scaling the environment size $W$ and the number of tasks $T_t$ to maintain constant robot density $(\lambda_N = N/{W^2})$ and constant task density $(\lambda_{T_t} = T_t/{W^2} =0.1)$. To ensure $T_t$ is time-invariant, we focused on the instant respawn scenario, using the M2 task setting (see Table \ref{tab:gen_levels}). Multiple robot densities were evaluated: $\lambda_N \in \{0.05, 0.1, 0.15, 0.2 \}$. 

\begin{figure}
    \centering
    \begin{subfigure}[] {.65\linewidth}
    \includegraphics[trim={.1in 0 .05in 0}, clip, width=1\linewidth]{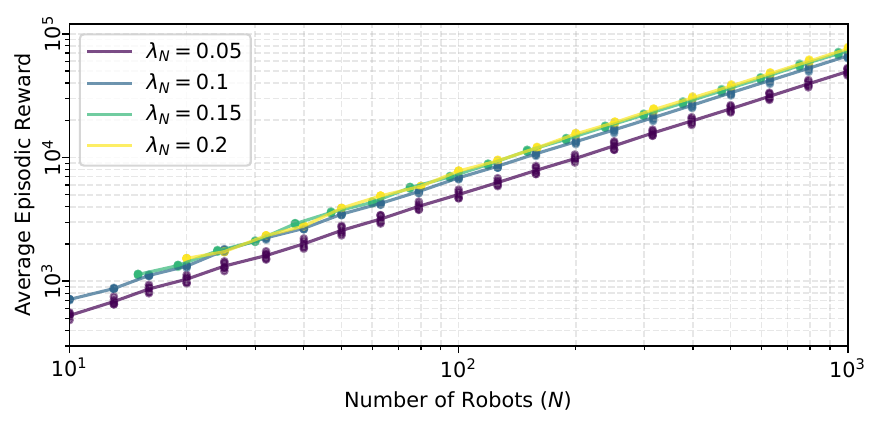}
    \caption{}
    \label{fig:scalability_plot}
    \end{subfigure}
    \hfill
    \begin{subfigure}[] {.315\linewidth}
    \includegraphics[trim={.05in 0 .1in 0}, clip, width=1\linewidth]{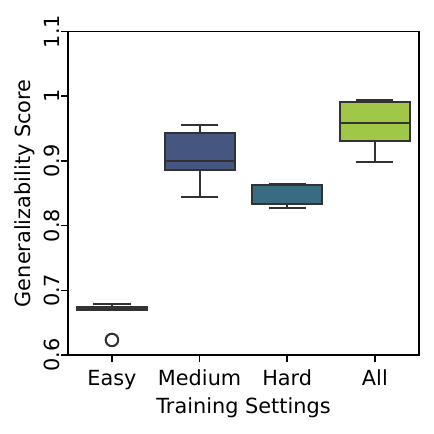} 
    \caption{}
    \label{fig:gen_boxplot}
    \end{subfigure}
    \caption{Scalability (a) and generalizability (b) results.}    
    \vspace{-1.0em}
\end{figure}

Fig. \ref{fig:scalability_plot} illustrates the performance of trained policies as a function of $N \in [10,1000]$. We observe that the policy's performance scales linearly with $N$ for all robot densities evaluated. These results validate the scalability of our method, demonstrating its effectiveness in managing larger numbers of robots for task allocation while maintaining constant robot and task densities.

To evaluate the generalizability of our method across environments with varying task settings, we train and evaluate our model on a diverse set of environments, each characterized by different task distributions and levels. 
As detailed in Table \ref{tab:gen_levels}, we designed nine task settings, categorized as easy, medium, or hard, each defined by a specific combination of task levels and quantities. 

To evaluate the performance of trained policies across different task settings, we introduce the \textit{generalizability score}. This score is calculated as the average of normalized rewards achieved by each trained policy within a specific category. The normalized reward is defined as the policy's average reward in a given setting, scaled by the maximum attainable reward. The maximum reward is estimated based on the average reward achieved by a policy trained specifically for that setting. Consequently, a policy with strong generalizability will achieve the score closer to 1.

Fig.~\ref{fig:gen_boxplot} shows the generalizability scores of policies trained in easy, medium, and hard task settings, as well as in a combined setting encompassing all three. The results demonstrate that the policy trained across all settings achieves better and more consistent generalization. This policy performs close to the highest performance attainable by setting-specific policies. These results indicate that our method enables trained policies to allow robots to achieve near-optimal performance across environments with varying task-level distributions when trained in diverse environment settings.

\section{Conclusions}
\label{sec:conclusions}

We introduced a learning-based framework to address decentralized dynamic coalition formation. By leveraging the \gls{mappo} algorithm and incorporating spatial action maps, motion planning, task allocation revision, and intention sharing, our model demonstrated significant performance improvements over existing approaches. 
Future directions include conducting experiments with physical multi-robot systems, such as multiple mobile manipulators collaboratively transporting heavy objects, where sensor noise and communication delays may substantially impact the performance of the proposed framework. Additionally, we plan to explore fine-tuning our learning-based framework using data collected from these real-world experiments to further optimize its performance.


\appendix
\textit{Modifications to PCFA:} 
We implemented \gls{pcfa} and integrated it with \gymrta\ for the performance comparison. To ensure a fair comparison with our method, the following modifications are applied:
\begin{enumerate*}[label=(\roman*)]
    \item \textbf{Utility Function}:
    The utility function $s_k$, associated with the $k$-th task, was adjusted to align with the reward function used in our framework:
    $s_k(t_{\text{ETA}}) = l_k^2 \, e^{-2(t_{\text{ETA}}+t_w-t_k^0)},$
    where $l_k$ is the task level, $t_{\text{ETA}}$ is the estimated time of arrival, $t_w$ is the wait time, and $t_{k}^0$ is the task's initial spawn time.

    \item \textbf{Partial Observability}: In \gls{pcfa}, the utility function for a task is set to zero if the task lies outside the robot's view range $R_{\text{view}}$, preventing robots from forming coalitions for tasks they cannot observe. Additionally, the communication graph is updated at each time step, with links existing only between robots that are within the communication range $R_{\text{comm}}$.
    
    \item \textbf{Task Assignment Revision}: To align with our method, robots are allowed to revise their task assignments at every time step. This is achieved by recomputing all four stages of \gls{pcfa} at every time step. While this modification increases communication overhead compared to the original \gls{pcfa}, it allows a fair comparison in our simulation study.
\end{enumerate*}

\balance
\bibliographystyle{ieeetr}
\bibliography{Bibliography}

\end{document}